# On true empty category

Qilin Tian    Zhejiang International Studies University[1]

**Abstract:** According to Chomsky (1981, 1986), empty categories consist of PRO, pro, trace, and variable. However, some empty object positions seem to be incompatible with extant empty categories. Given this, Li (2007a, 2007b, 2014) and Li & Wei (2014) raise the true empty category hypothesis, which holds that true empty category is only an empty position with category and Case features. As a last resort option, it is used mainly to meet the subcatgorization of a verb. This assumption is ingenious, and if proved to be true, it will exert a great impact on the study of UG. In this paper, we evaluate their evidence from topicalization and demonstrate that it can be accounted for without invoking true empty category.

**Key words:** true empty category, empty topic, pro

## 1. Introduction

In verbal communication, the intended meaning is delivered with sounds. However, it is not the case that there is always a strict correspondence between sounds and meanings. Sometimes, we will assign a specific meaning to a position where no sound is available within a sentence. For example, no sound is audible after the verb *xihuan* 'like' in (1). But we still know that what I like is this book rather than others.

(1) zhe-ben    shu,    wo feichang xihuan_____.
    this-CL    book    I    very    like.
    'As for this book, I like it very much.'

In the literature, the constituent that has no sound is normally named as empty category. According to Chomsky (1981, 1986), empty categories consist of PRO, trace, variable and pro. Huang (1984) and Huang, Li & Li (2009) argue that the distribution of pro and PRO is governed by the Generalized Control Rule (henceforth, GCR), shown in (2).

(2) Coindex an empty pronominal with the closest nominal element.        (Huang 1984:552)

In accordance with GCR, the empty object cannot be a Pro because if it is a Pro, it must be coindexed with the subject, the closest c-commanding nominal element. However, in accordance with Binding Condition B, it must have disjoint reference with the subject. This contradictory requirement indicates that it cannot be a Pro. It can only be a variable if it is coindexed with a constituent in an A'-position.

Nevertheless, it is pointed out by Li (2007a, 2007b, 2014) that the object position cannot be a variable under certain conditions. For example, the empty object in (3a) is in an island. Owing to island constraints, it cannot be a variable left the movement *zhege laoshi* 'this teacher' to the topic position. And the empty object in (3b) cannot be a variable either, because the elided constituent is indefinite and an indefinite constituent cannot move to the topic position.

(3) a. zhe-ge   laoshi$_i$ hen   hao,   wo mei kandao-guo [[e$_j$ bu   xihuan e$_i$ de] xuesheng$_j$].
       this-CL teacher very good    I   not see-ASP        not   like        DE student
       'This teacher$_2$ is very good. I have not seen students$_1$ who e$_1$ do not like (him$_2$).'

      b. ta kandao-le  yi-ge      nanhai; wo ye    kandao-le  (yi-ge nanhai).
         he see-ASP   one-CL   boy     I also   see-ASP    one-CL boy
         'He saw a book; I also saw (a boy).'

Given this, she raises the true empty category hypothesis, which assumes that true empty category (TEC henceforth) is not a novel kind of empty category like PRO and pro. Instead, it is just an empty position with only Case and category features, and it is a last resort option, used only when the existing empty category is not available to a particular empty argument position. In other words, only when no item can be taken from the lexicon to fill a position can we use TEC to meet the subcategorization of a verb. At LF, its interpretation observes the following rule.

(4)   A TEC is replaced at LF by what is available from the discourse context.

Li (2007a, 2007b, and 2014) and Li & Wei (2014) also find some evidence in support of this hypothesis, which includes subject/object asymmetry, topicalization constraint, and argument ellipsis. No doubt, their researches are of great significance because they discover some interesting facts. Furthermore, if proved to the true, TEC can deepen our understanding of the nature of empty categories, which will exert an important impact on the study of UG. In this paper, we evaluate their evidence from topicalization constraints and demonstrate that it can be accounted for without invoking TEC. Specifically, it will be argued that there is a base-generated covert topic, which can license itself by binding a pro in the comment clause. In addition to being supported by identity topic constructions in Chinese, this hypothesis can explain the reason why the empty object can coindex with a constituent outside an island, without resorting to TEC.

## 2.   Topicalization constraints

Li (2007a, 2007b, 2014) uses two kinds of phenomena related to topicalization to support the TEC. One is island constraints in topicalization, and the other is the constraint on indefinite topic. In this section we examine her evidence for each in turn.

To begin with, please see island constraints in topicalization. Li discovers that the empty object can be coindexed with an element outside an island, as (5) shows. She argues that this empty object cannot be a variable because overt movement of the object to the topic position across an island is ungrammatical, as is shown in (6).

(5)   ta$_i$ suiran  hen keqi,    keshi    ye    hen ai   mianzi.  ruguo
      he though  very polite   but      also   very love face    if
      [[nimen   piping    e$_i$  de taidu]    bu-hao],  ta   shi  bu  hui    ting      de ].
       you        criticize        DE attitude  not-good  he   be    not  will  listen   DE
     'Even though he is polite; he minds 'face' matters. If your attitude is not
      good criticizing (him), he will not listen.'

(6)  *ta$_i$      suiran   hen keqi,   keshi    ye    hen ai   mianzi.  ruguo ta$_i$
      he          though   very polite   but      also   very love face    if      he
      [[nimen  piping    e$_i$  de  taidu]   bu-hao],  ta   shi  bu  hui    ting      de ].
       you      criticize      DE attitude  not-good  he   be    not  will  listen   DE
     'Even though he is polite; he minds 'face' matters. If your attitude is not good criticizing (him), he will not listen.'

Based on this kind of facts, she argues that the empty object in (5) can only be a TEC. In our opinion, these facts can be explained without resorting to TEC. Concretely speaking, we propose that (5) and (6) are different in acceptability because while there is a covert topic base-generated in Spec-CP in (5), the overt topic in (6) is derived from movement. In other words, covert topic and overt topics are derived in different ways. Our assumption can not only account for the above fact, but also uncover the relationship between the topic construction and other constructions.

Before offering a detailed account for the above fact, we need to clarify and differentiate some concepts. The covert topic and the overt topic only refer to the phrase in Spec-CP with [+Topic] feature. Since [+Topic] feature has been incorporated into the syntactic system (Chomsky 1995), the covert topic and the overt topic are syntactic categories/syntactic topics. They are different from the discourse topic, which is a concept at the discourse level. Both the grammatical subject and the phrase in Spec-CP can serve as discourse topic.

Now let us consider the overt topic first. We assume that the overt topic (namely, the overt phrase in Spec-CP with [Topic] feature) is derived by movement if it is co-indexed with an empty position in the comment. This approach to topic constructions is supported by island constraints. For instance, the sentences in (7) are unacceptable because movement of the topic violates island constraints (see Huang 1982, Li 1990, Tian et al. 2024, among others).

(7) a. *zhe-ge ren$_i$,   Lisi xihuan e$_i$ de chuanyan   hen   duo.
      this-CL person   Lisi like    DE rumor       very   many
      'As for this person, there are many rumors saying that Lisi likes her.'
    b. *Zhangsan$_i$, guanyu e$_i$   changge shengyin de   baodao bu   duo.
       Zhangsan    about     singing voice   DE    report not   many
       'As for Zhangsan, there are few reports about his singing voice.'
    c. *Zhangsan$_i$, Lisi tiaowu de   yangzi he  e$_i$ changge de    shengying dou bu hao.
       Zhangsan   Lisi dancing DE matter  and    singing DE    voice   both not good
       'As for Zhangsan, both his singing voice and Lisi's dancing manner are not good'

(8a) lend support to our assumption from another perspective. Since the anaphor needs to be bound, *taziji de haizi* 'his child' must be originally generated in the object position of the predicate. (8b) demonstrates this point further. The ungrammaticality of this sentence suggests that island constraint is violated by movement of *taziji de haizi* 'his child' to the topic position.

(8) a. taziji   de   haizi, Zhangsan dou   bu   shede   gei   100 yuanqian.
       himself  DE   child Zhangsan even   not   willing   give 100 Yuan
       'Even if it is his son, Zhangsan is not willing to give him 100 Yuan.'
    b. *taziji$_i$   de   haizi, wo   renshi   gei-le   100   yuan de   na-ge    ren$_i$.
       himself    DE child  I    know    give-ASP 100 Yuan  DE that-CL   person
       'I know the person$_i$ who gave his$_i$ son 100 Yuan.'

Let us see (6) again. The topic 'ta' *him* is overt in the clause *ta nimen piping de taidu buhao* 'your attitude is not good criticizing (him)'. Therefore, it should be derived by movement. Owing to violation of the island constraint, (6) is expected to be ungrammatical.

However, different from (6), there is no overt topic in the second sentence of (5)[2]. Therefore, the grammaticality of this sentence is nothing unusual. It can be assumed that in (5) there is a covert topic (see Huang 1984), which is co-indexed with discourse topic *ta* 'he' (see Ai 2006: 133 for the coindexation between the covert topic and the discourse topic). When the covert topic is marked overtly, (5) will be like (9).

---

[2] There is no overt topic in (5)/(9). *Ta* 'he' is a grammatical subject as well as a discourse topic.

(9) ta<sub>i</sub> suiran hen keqi,    keshi   ye  hen ai   mianzi.  ruguo **[[Top Ø<sub>i</sub>]**
    he   though very polite   but    also very love face     if
    [nimen   piping   e<sub>i</sub>   de taidu]   bu-hao], ta shi bu hui ting   de ].
     you    criticize    DE attitude  not-good  he be  not will listen DE
    'Even though he is polite; he minds 'face' matters. If your attitude is not
     good criticizing (him), he will not listen.'

Distinct from Li (2014), who follows Huang (1984) in arguing that covert topics are also derived by movement, we propose that covert topics can be based-generated, which is licensed by be co-indexing with a pro in the comment clause[3].

Our assumption is supported by empirical evidence. First, some facts suggest that base-generated covert topic should be available. In Chinese, the pre-verbal NP can serve both as a topic and as a subject. For example, as an experiencer, *Zhangsan* in (10) is the highest argument in terms of the thematic hierarchy. Therefore, it should be a subject. But it can also bind a gap in the following clause, forming a topic chain, which is one of the characteristics of the topic (Tsao 1979, 1990).

(10) Zhangsan<sub>i</sub>   xihuan   yingyu,  [e<sub>i</sub>] zuotian canjia-le       yingyu  yanjiang
     Zhangsan    like     English      yesterday participate-LE   English  speech
     bisai,   [e<sub>i</sub>] bingqie hai  de-le       jiang.
     contest       and   also   win-ASP      prize.
    'Zhangsan likes English. Yesterday he participated in an English speech contest, and won a prize. '

To explain the dual status of *Zhangsan* in (10), we had better assume that there is a base-generated covert topic, which is identical to *Zhangan*.

It cannot be said that the subject in (10) has moved to the topic position in that such a movement is infeasible under certain circumstances. For example, being a focus in (11), *Zhangsan* cannot move overtly to the topic position. But it can still form a topic chain with a gap in the following clause[4].

(11) shi    Zhangan<sub>i</sub> zuotian  canjia-le    zhe-ge    huiyi,        erqie
     Focus Zhangan  yesterday participate-LE this-CL  conference   furthermore
     __e<sub>i</sub>_huidao    jia,    hai bei   laopo    ma__ e<sub>i</sub> __le.
         return     home   still passive  wife    scold     LE
    'It is Zhangsan who participated in this conference yesterday and when back at home, he was scolded by his wife.'

Above, we assume that there is a covert topic that is identical to the subject or the object. This means that there will be two identical constituents in a sentence, one being the topic and the other being the subject or the object. Interestingly, we do find many sentences in which the topic is

---

[3] See Pan & Hu (2008) and Hu & Pan (2009) for the terms 'license'.

[4] Shi (1989) argues that a constituent can establish a topic chain even if it cannot move to the topic position. He holds that the topic chain is also a basic unit, which is like a S'. This, in our opinion, is also a support to our hypothesis. If we assume that there is a covert topic heading the topic chain, the status of topic chain as S' will not be a surprise. Furthermore, there are many cases in which the first topic in the topic chain is covert, and such topic can coindex with an overt constituent, as well as empty categories. This can be seen clearly in (i) and (21c).

(i) [e<sub>i</sub>] meiyou kandao Zhangsan, wo<sub>i</sub> bu hui  gaosu ni  zhenxiang, [e<sub>i</sub>] ye  bu  hui zou.
       not    see   Zhangsan   I  not will tell    you truth         also not will leave
   'Not seeing Zhangsan, I will not tell you the truth, and will not leave, either.'

identical to a constituent in the comment, as shown in (12). The existence of such topics, or identical topics in terms of Liu (2004), lends strong support to our hypothesis.

(12) a. shui    shui    jinzhang, dian    dian    jinzhang, (meiqi
       water   water   short     electricity electricity short   gas
       meiqi   jingzhang).
       gas     short
       'Speaking of water, it is short; speaking of electricity, it is short; and speaking of gas, it is short, too.'
   b. xiangyan    me,   wo   yiqian   ye   chou-guo    xiangxiang.
      tobacco    TOP   I    before   also smoke-ASP   tabacco
      'As for smoking, I did smoke before.'
   c. Zhangsan,   shui    shui    bu    he,    fan    fan    bu    chi.
      Zhangsan   water   water   not   drink  food   foot   not   eat
      'Speaking of water, Zhangsan does not drink; and speaking of food, Zhangsan does not eat.'

It can be said that normally the identical topic is covert, as in (10), and only when it is used for emphasis or contrast will it become overt, as in (12).

Our assumption is also theoretically supported that the base-generated covert topic is co-indexed with a pro. First, Huang (1982), Li (1990), Huang, Li & Li (2009) argue that overt topics in Chinese can be derived in two ways: base-generation and movement. The so-called Chinese style topics like *shuiguo* 'fruit' in (13a) are base-generated in Spec-CP, and the topic in (13b), which is related to a gap in the comment, is derived by movement (see Tian et al. 2024 for different views about sentences like (13a)).

(13) a. shuiguo, wo   xihuan   pingguo.
       fruit    I    like     apple
       'As for the fruit, I like apples.'
   b. zhe-ben    shu,    wo    feichang xihuan_____.
      this-CL   book    I     very     like
      'As for this book, I like it very much.'

Since the overt topic is likely to be base-generated in Spec-CP, it is theoretically possible for the covert topic to be base-generated in Spec-CP for the sake of symmetry.

Second, it has been argued by many scholars that overt dependency and covert dependency are established differently. For example, Turano (1998), among others, claims that overt dependency is derived by movement, but covert dependency is derived by binding. That is, no movement is involved in the latter. In Chinese we can find several kinds of constructions, in which covert dependency is established without movement. The following are some typical examples.

(14) a. zhe-ge    shihou,   shui   xie   de   shu   dou   bu   hao    mai.
       this-CL  moment    who    write DE   book  all   not  easy   sell
       'At this moment, no one's book is easy to sell.'
   b. ruguo shui   xie    de    shu    mai   chuqu-le, na   ta    jiu   yao   qingke.
      if    who   write  DE    book   sell  out-ASP  then he    then must  invite-guest
      'If someone sold his book, then he should pay the bill.'

    c. shui     mai de  shu,    shui kan.
       who     buy DE book   who read
       'The person who bought the book read it.'
    d. Zhangan   xihuan     shui mai de yifu?
       Zhangsan   like       who buy DE clothes
       '*Zhangsan like the clothes that who bought?'

In all these constructions, covert dependency is established without resorting to movement. Therefore, no island effect can be observed. Considering these, we can claim while overt topic (if coindexed with an empty position in the comment clause) is derived by movement, the covert dependency in the topic construction should be derived without movement.

Third, Reuland (2013) argues that the pronoun can be translated as a variable in logical syntax/at the C-I interface. If the empty object can be a pro, it should also be able to be translated as a variable at the C-I interface, which is bound by the covert topic in Spec-CP. Interestingly, Huang (1984) has raised similar assumption although we are different in that he argues that an overt topics can co-index with a pro in the subject position, which become a variable at LF, and we assume that only covert topic is able to be licensed this way. Anyway, the above similarity suggests that our assumption might be on the right track.

Besides, a lot of scholars like Tsai (1994) takes an unselective binding approach to wh's-in-situ shown in (14). If our assumption is feasible, we might uncover the connection between topic constructions, interrogative sentences, and 'bare' conditionals, etc. Consequently, we might be a step further towards opening a window to the mystery of UG.

Forth, the main evidence against the presence of a pro in object position comes from the combined effect of the GCR and Binding Condition B. However, it has been pointed out by Zhang (2016) and Ke (2023), among others, that GCR is problematic. One of Zhang's (2016) examples is listed below. In this sentence, the empty subject is coindexed with *Lili* rather than the closest higher subject *A-Bao*, contrary to the prediction of GCR.

(15) A-Gui xian   [Lili$_i$ tai aoman],   A-Bao xian   [pro$_i$ tai luosuo].
     A-Gui dislike Lili too arrogant    A-Bao dislike     too long-winded
     'A-Gui dislikes Lili's being too arrogant and A-Bao dislikes her being long-winded.'

In sum, it is demonstrated above that it is unsurprising that the empty object can coindex with a constituent outside an island in (5) because the empty object can co-index with a based-generated covert topic outside an island, which co-refer with *ta* 'he' in the first sentence. Since no grammatical constraint is violated, (5) is expected to be acceptable.

Now let us see the constraint on indefinite topic, which is used by Li (2014) as evidence in support TEC. Take (3b)(repeated as (16)) as an example. Li (2014) argues that since indefinite phrase cannot serve as a topic, the null object cannot be regarded as a variable bound by the null topic *yi-ge nanhai* 'a boy'. Since it fails to function as a variable, the null object should be a TEC.

(16) ta kandao-le yi-ge    nanhai; wo ye    kandao-le  (yi-ge   nanhai).
     he see-Asp  one-CL boy     I   also   see-ASP    one-CL boy
     'He saw a book; I also saw (a boy).'

We think that this evidence is not convincing, either. At the very beginning, a fact needs to be pointed out, which to our knowledge, has been rarely discussed. That is, the elided constituent in (16) needs not be identical to its antecedent in number, which can be seen more clearly in (17).

(17) a. wo    tou        kan-le     yi-ben    xiaoshuo, ta    ye     tou         kan-le.
        I     stealthily read-ASP  one-CL    novel     he    also  stealthily  read-ASP
     'I read a novel stealthily, and he also read novels stealthily.'

   b. wo       mai-le    yi-ben    shu,       ni mai-le ____ ma?
      I        buy-ASP  one-CL   book       you buy-ASP     Q
      'I bought a book. Have you bought books?'

For example, the second clause in (34a) only means that he also read novels stealthily. It does not specify the number of novels. Similarly, the null object in (16) should be interpreted as 'I also saw boys, with the number of boys unspecified'. Li's (2014) translation of (16) is just one possible interpretation. Given this, it can be said that the covert topic of the second conjunct in (16) should be *nanhai* 'boy' rather than *yige nanhai* 'one boy'. And then, it is possible to bind a variable in the object position. As a result, TEC becomes unmotivated.

As a matter of fact, with our covert topic hypothesis, the above phenomenon can be accounted for more reasonably. Before offering an account of this phenomenon, let us have a look at the derivational mechanism of ellipsis. Up to now, different analyses have been proposed to account for ellipsis. Generally speaking, the structural approach to ellipsis can be classified into two groups (see Merchant 2016, Murphy & Muller 2017): (i) those that claim that there are unpronounced syntactic structures in the ellipsis site (this is also named as PF deletion approach, see Merchant (2001)), and (ii) those that do not (Lobeck 1995). Both approaches have their own strengths as well as weaknesses. In this paper, we will adopt Baltin's (2012) assumption which is compatible with both analyses. Baltin (2012) assumes that ellipsis should take place in the narrow syntax rather than at PF (see also Murphy & Muller 2017, Park 2017). With his assumption, we can explain why there is a connectivity effect between the structure containing the ellipsis and the unpronounced constituents, such as case matching shown in (18).

(18) Er will    jemandem         schmeicheln, aber sie     wissen nicht, wem/      *wen.
     he wants  someone.DAT     flatter       but  they   know    not   who.DAT  who.ACC
     'He wants to flatter someone, but they don't know who.' (Merchant 2001:42)

It can be said that all the syntactic structures are built at the very beginning of the derivation, which leads to the case matching effect. And some parts of the merged structure are elided in the narrow syntax, which can explain why the elided constituent behaves like a pro at the interface (Lobeck 1995, Chung, Ladusaw & McCloskey 1995).

With the assumptions above, we can succeed in explaining (16) and (17). For example, after the object is elided in the narrow syntax in (17a), it will be similar to a pro at the interface, which is bound by a covert topic[5]. The covert topic in the second clause should be *xiaoshuo* 'novel'

---
[5] One point that needs to be clarified is that even though we label the object as a pro after it is elided, it is different from the overt pronoun. The following two sentences can illustrate this point.

  (i) Zhangsan zunzhong ziji   de   mama,   Lisi   ye   zunzhong.
      Zhangsan respect    self  DE  mother  Lisi  also  respect
      'Zhangsan$_i$ respect his mother, and Lisi$_j$ other respect his$_{i/j}$ mother.' (sloppy reading and strict reading)
  (ii) Zhangsan zunzhong ziji   de   mama,   Lisi   ye   zunzhong ta.
      Zhangsan respect    self  DE  mother  Lisi  also  respect   her
      'Zhangsan$_i$ respect his mother, and Lisi other respect his$_i$ mother.' (strict reading only)

As can be seen, while the elided structure allows both sloppy reading and strict reading, the overt pronoun only allows strictly reading. This difference may lie in that after the object is elided, it becomes an empty element which needs to be interpreted at the interface. It can be interpreted by copying the covert topic to the object position (just like the reconstruction operation), where the empty object can acquire both sloppy reading and strict rather than *yiben xiaoshuo* 'a novel' because the indefinite phrase cannot serve as a topic. As a result, the second clause of this sentence will mean that he also read a novel /novels.

## 3. Conclusion

Li (2007a, 2007b, 2014) and Li & Wei (2014) raise the TEC hypothesis, which holds that the empty object can be an empty position with only Case and category features. It is a last resort option. That is, it is used only when extant empty categories, such as PRO, pro, trace and variable, are not available. This hypothesis is very interesting. With such a hypothesis, Li uncovers many important facts. Furthermore, if proved to be true, TEC is also of great theoretical significance because it can enrich our understanding of the nature of empty categories. In this paper, we evaluate their evidence from topicalization and demonstrate that it can be accounted for without invoking TEC. With a covert topic base-generated in Spec-CP, we not only show that it is nothing special for the empty object to coindex with a constituent outside an island, but also bring to light the connection between the covert topic construction, identical topic constructions, and the wh-in-situ constructions. In the future, we will evaluate other evidence for TEC along the line above, demonstrating that the evidence can also be explained without TEC.

---

reading(see Baltin 2012). By contrast, the object in the second conjunct of (ii) is an overt pronoun, which cannot be interpreted by the LF copying operation, or the reconstruction operation. Therefore, it has only strict reading.

In this paper, for ease of illustration, we simply assume that after a constituent is elided, it is just like a pro, which is bound by a covert topic.